\documentclass[letterpaper, 10 pt, conference]{ieeeconf}  % Comment this line out if you need a4paper

\IEEEoverridecommandlockouts                              % This command is only needed if 
\usepackage{multicol}
\usepackage{multirow}
\usepackage{float} % Required for [t] option
\usepackage{siunitx} % for rounding
\usepackage{tablefootnote}
\usepackage{booktabs} % For cleaner table lines
\usepackage{svg}
\usepackage{graphicx}
\usepackage{subcaption}
\usepackage{amsmath}
\usepackage{gensymb}
\usepackage{amssymb} %for Real number symbol
\usepackage{relsize} % for larger symbols
\usepackage{cite} % citation numbers will automatically be sorted and ranged IEEE style.

\makeatletter
\let\NAT@parse\undefined
\makeatother
\usepackage[hidelinks]{hyperref}

\usepackage[capitalize]{cleveref}
\crefname{figure}{Fig.}{Figs.}
\Crefname{figure}{Figure}{Figures}
\crefname{table}{Table}{Tables}
\Crefname{table}{Table}{Tables}
\title{\LARGE \bf
Toward Robust LiDAR Semantic Segmentation for Real-World Deployment: Evaluation under Coarse Labels, Adverse Conditions, and Domain Shifts}

\author{Samir Abou Haidar$^{1}$, Alexandre Chariot$^{2}$, Mehdi Darouich$^{2}$, Cyril Joly$^{1}$ and Jean-Emmanuel Deschaud$^{1}$% <-this % stops a space
\thanks{$^{1}$The authors are with Mines Paris, PSL University, Centre for Robotics (CAOR), 75006 Paris, France
        {\tt\small first\_name.last\_name@minesparis.psl.eu}}%
\thanks{$^{2}$The authors are with Paris-Saclay University, CEA, List, F-91120, Palaiseau, France
        {\tt\small first\_name.last\_name@cea.fr}}%
}

\begin{document}

\maketitle
\thispagestyle{empty}
\pagestyle{empty}

\begin{abstract}
LiDAR-based semantic segmentation is a core perception module for autonomous vehicles and mobile robots. Despite the strong performance of recent state-of-the-art methods on standard benchmarks, existing evaluation protocols remain focused on clean, single-domain settings and fine-grained label taxonomies, leaving deployment readiness largely unassessed. Real-world systems must handle safety-critical label semantics, degraded sensing conditions, and cross-domain variability, yet no unified protocol currently addresses all three aspects together. In this paper, we propose a structured evaluation protocol that assesses the deployment readiness of LiDAR semantic segmentation models along three complementary dimensions: (i) coarse-label evaluation aligned with autonomous driving safety priorities, revealing how label granularity affects different methods; (ii) robustness under eight types of LiDAR corruptions designed to emulate real-world atmospheric, geometric, and sensor degradations; and (iii) domain generalization across datasets without adaptation. The evaluation includes inference speed measured on an embedded Jetson AGX Orin platform, directly reflecting deployment constraints. Our results show that fine-grained benchmark rankings do not always reflect safety-relevant performance, that all methods experience substantial degradation under corruptions with architecture-dependent robustness characteristics, and that current domain generalization remains insufficient for reliable deployment. These findings expose concrete gaps between benchmark performance and deployment readiness, and provide a reference protocol for more practically grounded evaluation of LiDAR semantic segmentation.
\end{abstract}
    
\section{INTRODUCTION}
\label{sec:introduction}

Autonomous vehicles and mobile robots rely extensively on LiDAR sensors to capture 3D point cloud representations of their surroundings. These sensors provide accurate geometric information that enables perception systems to understand complex outdoor environments. Within this perception stack, LiDAR semantic segmentation plays a central role by assigning a semantic label to every point in the 3D point cloud, allowing the system to distinguish between relevant scene elements such as vehicles, pedestrians, roads, and buildings. This semantic understanding is essential for downstream tasks including object detection, scene interpretation, and decision-making, which are critical for safe autonomous navigation.

In recent years, numerous deep learning approaches have been proposed for LiDAR semantic segmentation, achieving strong performance on widely used benchmarks such as SemanticKITTI~\cite{behley2019semantickitti} and nuScenes~\cite{caesar2020nuscenes}. Despite these advances, the evaluation of LiDAR semantic segmentation methods remains largely centered on standard train/val/test splits of these datasets, using fine-grained class taxonomies and clean data. While these benchmarks have been crucial in driving progress, they represent only a first step toward real-world deployment. Models intended to run on autonomous robotic platforms must satisfy a far broader set of requirements: they must perform reliably under safety-critical label semantics, remain functional under sensor degradation, generalize across previously unseen environments, and deliver these capabilities within the computational and latency constraints of embedded platforms used in real-world autonomous systems.

Despite significant progress in the literature, no existing evaluation protocol jointly assesses semantic relevance, robustness to sensor degradations, cross-domain generalization, and computational efficiency. Standard benchmarks~\cite{behley2019semantickitti, caesar2020nuscenes} focus on in-distribution accuracy using fine-grained label taxonomies. The Robo3D benchmark~\cite{kong2023robo3d} evaluates robustness to sensor corruptions but does not consider safety-oriented label semantics or cross-domain generalization. The COLA framework~\cite{sanchezcola2025} introduces coarse-label evaluation and cross-dataset comparison but does not assess corruption robustness or computational efficiency. Meanwhile, domain generalization studies~\cite{sanchez2023domaingeneralization, sanchez20253dlabelprop} investigate cross-domain transfer but typically focus on a single evaluation axis. As a result, current evaluation practices provide only a fragmented view of deployment readiness, making it difficult to assess how models trade off accuracy, robustness, generalization, and runtime performance in realistic operating conditions.

To address this gap, we propose a unified evaluation protocol structured around three complementary pillars that capture key requirements for autonomous driving perception:

\textbf{Coarse-label evaluation} assesses semantic segmentation performance under label groupings aligned with autonomous driving safety priorities. Beyond reporting results with coarser taxonomies, we analyze how models perform with fine- and coarse-grained labels, providing insight into the extent to which benchmark outcomes are influenced by annotation granularity. To this end, we adopt the COLA framework~\cite{sanchezcola2025}, which defines a safety-oriented taxonomy that groups semantically related classes according to their relevance for autonomous driving.

\textbf{Robustness evaluation} measures model performance under eight types of synthetic LiDAR corruptions designed to emulate realistic sensing degradations. These include atmospheric effects such as fog, rain, and snow; geometric perturbations such as motion blur and missing beams; and sensor-induced artifacts including crosstalk, incomplete echoes, and cross-sensor variations. Corruptions are analyzed both individually and by category, enabling systematic comparison of robustness characteristics across architectures. Inference speed on Jetson AGX Orin embedded hardware is reported alongside robustness results to quantify the trade-offs between robustness and computational efficiency.

\textbf{Domain generalization evaluation} assesses the ability of models trained on one dataset to operate in a previously unseen target domain without any adaptation. Specifically, models are trained on SemanticKITTI~\cite{behley2019semantickitti} and nuScenes~\cite{caesar2020nuscenes} and evaluated directly on ParisLuco3D~\cite{sanchez2024parisluco3d} without any adaptation. This setting measures generalization across different geographic environments, acquisition conditions, and dataset distributions. By comparing performance on the source and target domains, we quantify the extent of cross-domain degradation and assess the robustness of the methods to distribution shift.

Using this protocol, we evaluate a representative set of state-of-the-art methods spanning point-based, projection-based, sparse convolution-based, and fusion-based architectures. All models are retrained from scratch under consistent training settings to ensure fair comparison. The resulting analysis provides a comprehensive characterization of the strengths and limitations of current LiDAR semantic segmentation approaches and reveals the trade-offs between semantic accuracy, robustness, generalization capability, and computational efficiency. These insights contribute toward a more realistic assessment of model suitability for autonomous driving applications.

\section{RELATED WORK}
\label{sec:relatedwork}

\subsection{LiDAR Semantic Segmentation Architectures}

LiDAR semantic segmentation methods are commonly categorized by their input representation. \textbf{Point-based methods} operate directly on raw 3D points, preserving full geometric information. WaffleIron~\cite{puy2023waffleiron} and PTv3~\cite{wu2024ptv3} introduced local feature aggregation techniques to improve accuracy, but their high computational costs limit practical use in real-time scenarios. \textbf{Projection-based methods} such as SalsaNext~\cite{cortinhal2020salsanext}, CENet~\cite{cheng2022cenet}, and FRNet~\cite{xu2025frnet} project the point cloud onto 2D range images and often apply CNN-based processing, offering high throughput at the cost of some spatial accuracy. \textbf{Sparse convolution-based methods} such as Minkowski~\cite{choy2019minkowski} and Cylinder3D~\cite{zhu2021cylindrical} focus computations on occupied or non-empty voxels, achieving a good balance between accuracy and efficiency. \textbf{Fusion-based methods} such as SPVCNN~\cite{tang2020spvnas}, and HARP-NeXt~\cite{abouhaidar2025harpnext} combine multiple representations to leverage complementary information and enhance the accuracy.

\subsection{Evaluation Protocols for LiDAR Semantic Segmentation}

Evaluation of LiDAR semantic segmentation has traditionally followed the standard dataset partitioning (train/val/test) of individual benchmarks, with SemanticKITTI~\cite{behley2019semantickitti} and nuScenes~\cite{caesar2020nuscenes} serving as the primary references. These protocols measure in-distribution accuracy using fine-grained class taxonomies and clean point cloud data, which does not fully reflect the challenges of real world deployment.

Several works have proposed alternative evaluation strategies. The Robo3D benchmark~\cite{kong2023robo3d} systematically evaluates robustness to LiDAR corruptions, including weather-, geometry-, and sensor-related degradations, and demonstrates that many state-of-the-art models suffer substantial performance degradation under realistic perturbations. However, it focuses solely on robustness and does not consider label semantics, domain generalization, or inference speed. The COLA framework~\cite{sanchezcola2025} addresses the label granularity problem by proposing coarse semantic groupings aligned with autonomous driving safety requirements, enabling consistent comparison across datasets. However, it does not assess robustness or runtime constraints. Additionally, domain generalization for LiDAR semantic segmentation remains relatively underexplored compared with 2D vision~\cite{li2018learningtogeneralize, li2019episodictraining} and 3D domain adaptation~\cite{yi2021domainadaptation}. Existing studies analyze the impact of sensor, appearance, and scene shifts~\cite{sanchez2023domaingeneralization, sanchez20253dlabelprop} and investigate methods such as style augmentation~\cite{jin2022stylenormalization}, domain randomization~\cite{langer2020domaintransfer}, and episodic training~\cite{li2019episodictraining}.

Unlike prior evaluation protocols, which typically focus on a single aspect of model performance, our protocol integrates semantic consistency, robustness, domain generalization, and computational efficiency into a unified assessment framework. By evaluating these dimensions under consistent experimental conditions, it provides a more comprehensive assessment of model suitability for real-world autonomous driving deployment.

\section{EVALUATION PROTOCOL}
\label{sec:methodology}

We propose a structured evaluation protocol to assess the deployment readiness of LiDAR semantic segmentation models. The protocol is organized around three complementary dimensions that characterize model behavior under conditions relevant to real-world autonomous driving: safety-oriented label semantics, adverse sensing conditions, and cross-domain transfer. In addition, computational efficiency is assessed through inference speed on both an NVIDIA RTX 4090 GPU and a Jetson AGX Orin embedded platform. All evaluated models are retrained from scratch under consistent training conditions to ensure a fair comparison.

\subsection{Coarse-label evaluation for safety assessment}
\label{subsec:coarselabelsemseg}

Standard benchmarks such as nuScenes~\cite{caesar2020nuscenes} and SemanticKITTI~\cite{behley2019semantickitti} evaluate models using fine-grained class taxonomies with 16 and 19 classes, respectively.
\begin{figure}[h]
  \centering
  \begin{subfigure}[b]{0.485\textwidth}
    \centering
    \includegraphics[width=\textwidth]{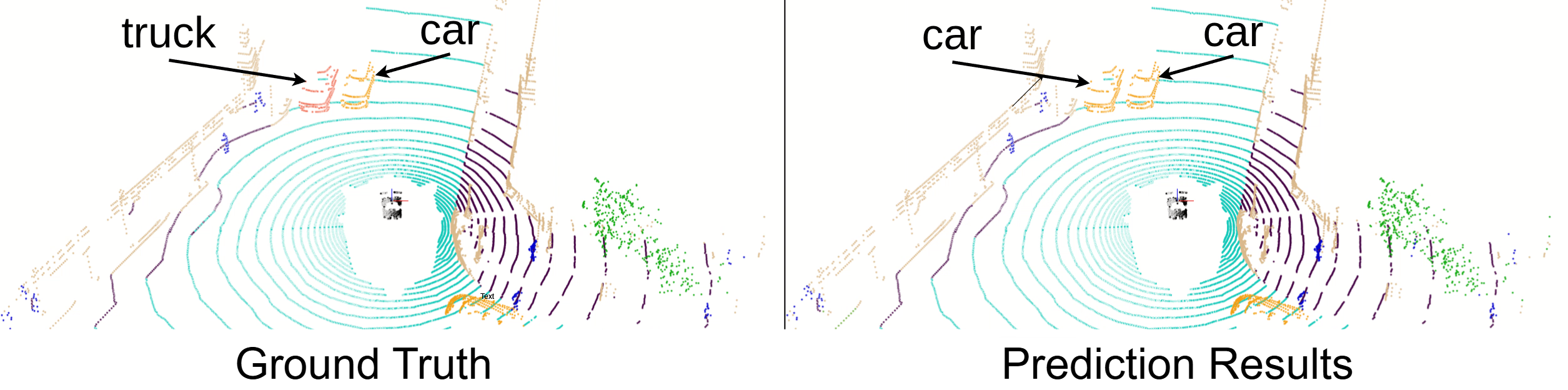}
    \caption{Car and truck confusion in nuScenes~\cite{caesar2020nuscenes}.}
    \label{fig:colanuscenes_car}
  \end{subfigure}

  \vspace{0.2em}

  \begin{subfigure}[b]{0.485\textwidth}
    \centering
    \includegraphics[width=\textwidth]{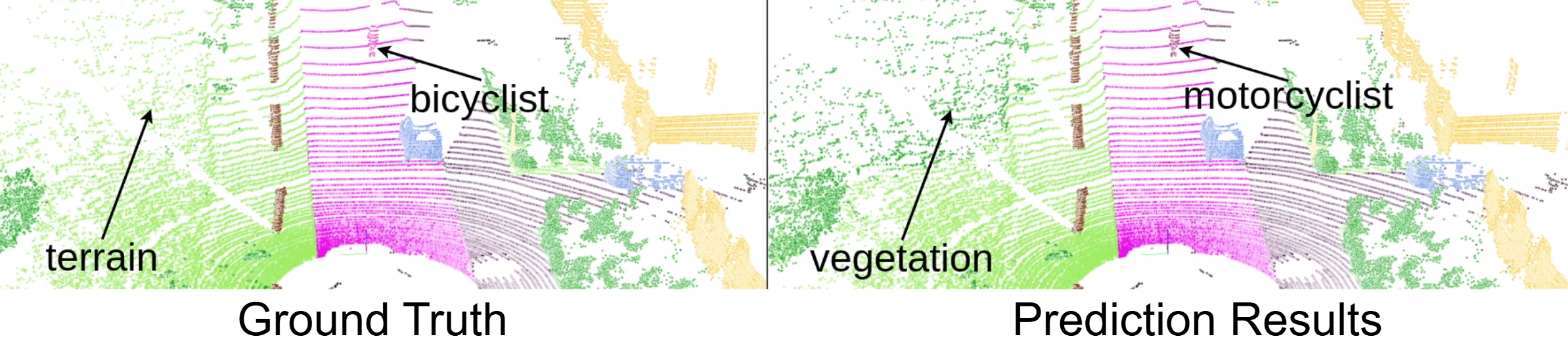}
    \caption{Bicyclist and motorcyclist confusion in SemanticKITTI~\cite{behley2019semantickitti}.}
    \label{fig:colasemkitti_bike}
  \end{subfigure}

  \caption{Examples of fine-grained semantic misclassifications arising from geometric and contextual ambiguity.}
  \label{fig:finegrainedmissclassifications}
\end{figure}
While these annotations enable detailed scene understanding, fine-grained labels can lead to overly strict evaluations that penalize semantically reasonable predictions from a safety and operational perspective. In particular, confusions between functionally similar classes (e.g., car vs.\ truck, or bicyclist vs.\ motorcyclist as shown in \cref{fig:finegrainedmissclassifications}) may substantially reduce reported mIoU despite having limited impact on many autonomous driving decisions. For example, confusing a car with a truck would often result in similar avoidance or braking behavior, even though it is counted as a complete error under conventional fine-grained evaluation.

To address this limitation, we adopt the COarse-LAbel (COLA) framework~\cite{sanchezcola2025}, which groups dataset-specific semantic labels according to their functional role in autonomous navigation. Beyond reporting coarse-label performance, we analyze how model performance changes when transitioning from fine- to coarse-grained taxonomies.

Under this framework, semantically and functionally related classes are merged into seven coarse categories: \textit{Driveable Ground}, \textit{Other Ground}, \textit{Structure}, \textit{Vehicle}, \textit{Nature}, \textit{Living Being}, and \textit{Object}. These categories correspond to the semantic distinctions most relevant to autonomous navigation, including identifying the driveable surface, detecting dynamic agents, and recognizing static structures and obstacles. For example, all vehicle-related classes (e.g., car, bus, truck, construction vehicle) are merged under \textit{Vehicle}, while vulnerable road users (pedestrians, bicyclists, and motorcyclists) are grouped under \textit{Living Being}. Confusions \textit{within} a coarse category (e.g., car vs.\ truck) do not affect mIoU$^{CO}_{LA}$, whereas confusions \textit{across} categories (e.g., a vehicle point predicted as road) remain penalized, reflecting safety-relevant failures. In addition to segmentation performance, inference speed on an NVIDIA RTX 4090 GPU is reported to provide a reference measure of computational efficiency.

\subsection{Robustness under real-world LiDAR corruptions}
\label{subsec:robustness}

In real-world deployments, LiDAR sensors operate under a wide range of conditions that degrade point cloud quality in ways clean benchmarks do not capture. To evaluate robustness systematically, we adopt the corrupted datasets \textit{nuScenes-C} and \textit{SemanticKITTI-C} from the Robo3D benchmark~\cite{kong2023robo3d}. Models are trained on clean datasets and evaluated directly on corrupted versions without fine-tuning, ensuring the evaluation reflects out-of-distribution robustness.

Robo3D introduces eight corruption types organized into three semantically meaningful groups:
\begin{itemize}
    \item \textbf{Atmospheric} (Fog, Wet Ground, Snow): environmental conditions affecting laser propagation.
    \item \textbf{Geometric} (Motion Blur, Beam Missing): motion dynamics and beam dropout effects.
    \item \textbf{Sensor} (LiDAR Crosstalk, Incomplete Echo, Cross-Sensor): internal sensor characteristics or hardware differences.
\end{itemize}

We adopt the mean Corruption Error (mCE) as the primary robustness metric~\cite{dan2019benchmarkingrobustness, kong2023robo3d}:
\begin{equation}
\text{CE}_i = \frac{\sum_{l=1}^{3} (1 - \text{Acc}_{i,l})}{\sum_{l=1}^{3} (1 - \text{Acc}^{\text{baseline}}_{i,l})}, \quad
\text{mCE} = \frac{1}{N} \sum_{i=1}^{N} \text{CE}_i ,
\end{equation}
where $\text{Acc}_{i,l}$ is the mIoU for corruption type $i$ at severity $l$, $N=8$, and MinkowskiNet~\cite{choy2019minkowski} is the normalization baseline.

We define real-time operation as inference under 100\,ms per scan on SemanticKITTI ($>$10 FPS) and under 50\,ms per scan on nuScenes ($>$20 FPS) on the Jetson AGX Orin. These thresholds correspond to the acquisition frequencies of the datasets' LiDAR sensors, namely the 10 Hz Velodyne HDL-64E used in SemanticKITTI~\cite{behley2019semantickitti} and the 20 Hz Velodyne HDL-32E used in nuScenes~\cite{caesar2020nuscenes}. Real-time operation is therefore defined with respect to single-scan inference, ensuring that the processing of one LiDAR sweep is completed before the arrival of the next scan.

\subsection{Domain generalization}
\label{subsec:domaingeneralization}

Domain generalization evaluates a model's ability to operate in previously unseen environments without any adaptation or access to target-domain data during training. In the context of LiDAR semantic segmentation, this requires robustness to shifts in sensor characteristics, environmental appearance, and scene composition.

We evaluate models under a \textbf{single-source} to \textbf{single-target} protocol: a model trained on a source dataset is evaluated directly on an unseen target dataset without adaptation. We use nuScenes~\cite{caesar2020nuscenes} and SemanticKITTI~\cite{behley2019semantickitti} as source datasets, and ParisLuco3D~\cite{sanchez2024parisluco3d} as the target dataset. ParisLuco3D was specifically designed for cross-dataset evaluation and differs from both training datasets in geographic environment, acquisition conditions, and scene composition.

Evaluation is performed using the native fine-grained label taxonomy of each source dataset. Since no label grouping or adaptation is applied, the resulting per-class IoU scores reflect the full impact of domain shift on detailed semantic segmentation performance.

\section{RESULTS AND DISCUSSIONS}
\label{sec:experimentalresults}

We evaluate a representative set of methods spanning the four main architectural families for LiDAR semantic segmentation: point-based (KPConv~\cite{thomas2019kpconv}, WaffleIron~\cite{puy2023waffleiron}, and PTv3~\cite{wu2024ptv3}), projection-based (SalsaNext~\cite{cortinhal2020salsanext}, CENet~\cite{cheng2022cenet}, and FRNet~\cite{xu2025frnet}), sparse convolution-based (Minkowski~\cite{choy2019minkowski} and Cylinder3D~\cite{zhu2021cylindrical}), and fusion-based (SPVCNN~\cite{tang2020spvnas}, and HARP-NeXt~\cite{abouhaidar2025harpnext}). These approaches were selected as both established baselines and well-performing methods on major benchmarks including SemanticKITTI~\cite{behley2019semantickitti} and nuScenes~\cite{caesar2020nuscenes}. This diversity ensures that the benchmark conclusions generalize across different architectural paradigms. 

To ensure a fair comparison, all models are reproduced and retrained from scratch under a unified training protocol (see \cref{tab:trainingprotocol}) on a single NVIDIA GeForce RTX 4090 GPU. We apply standard point cloud augmentations, including random rotation around the $z$-axis, random flipping along the $x$ and $y$ axes, and random scaling. To further address class imbalance in SemanticKITTI~\cite{behley2019semantickitti}, we employ instance-level augmentation strategies, including PolarMix~\cite{xiao2022polarmix} and Instance CutMix~\cite{puy2023waffleiron}. These methods augment scenes by inserting transformed instances of rare classes (bicycle, motorcycle, person, bicyclist, and other-vehicle) into randomly selected locations within the point cloud.

\begin{table}[h]
    \centering
    \begin{tabular}{c c c}
    \toprule
    \textbf{Config} & \textbf{SemanticKITTI~\cite{behley2019semantickitti}} & \textbf{nuScenes~\cite{caesar2020nuscenes}} \\ 
    \midrule
    Optimizer & AdamW & AdamW \\
    Scheduler & WarmupCosine & WarmupCosine \\
    \multirow{3}{*}{Criteria} & CrossEntropy & CrossEntropy \\
     & Lovász-Softmax~\cite{berman2018lovaszsoftmaxloss} & Lovász-Softmax~\cite{berman2018lovaszsoftmaxloss} \\
     & Boundary$^*$~\cite{bokhovkin2019boundaryloss} & Boundary$^*$~\cite{bokhovkin2019boundaryloss} \\
    Max learning rate & 1e-3 & 1e-3 \\
    Min learning rate & 1e-5 & 1e-5 \\
    Weight decay & 3e-3 & 3e-3 \\
    Batch size & 2 & 4 \\
    Warmup epochs & 4 & 4 \\
    Epochs & 80 & 100 \\
    \bottomrule
    \multicolumn{3}{l}{$^*$Applied to projection-based methods only.}
    \end{tabular}
    \caption{Networks unified training protocol.}
    \label{tab:trainingprotocol}
\end{table}

Inference speed is measured on both a high-end RTX 4090 GPU (reference high-throughput setting) and an embedded NVIDIA Jetson AGX Orin (2048-core Ampere GPU, 12-core Arm Cortex-A78AE CPU, 64 GB memory) to reflect real-world deployment constraints. Reported FPS values in \cref{tab:coarseperformancenuscenes,tab:coarseperformancesemantickitti,tab:robustness} correspond to end-to-end runtime, including both data pre-processing and network inference.

\subsection{Coarse-label evaluation}
\label{subsec:colaresults}

In this section, we evaluate LiDAR semantic segmentation models under the Coarse Label (COLA) annotation~\cite{sanchezcola2025}, which aggregates fine-grained classes into broader semantic categories. Such evaluation better reflects real-world autonomous driving perception tasks, where identifying high-level classes (e.g., \textit{vehicle} or \textit{living-being}) is often more relevant than distinguishing between individual subclasses (e.g., \textit{car} or \textit{truck}). COLA therefore provides complementary insights into semantic scene understanding beyond conventional fine-grained evaluation.

As reported \cref{tab:coarseperformancenuscenes} and \cref{tab:coarseperformancesemantickitti}, all evaluated methods exhibit a consistent improvement when transitioning from fine-grained mIoU$^{FG}$ to the coarse-level mIoU$^{CO}_{LA}$. This trend indicates that errors at the fine semantic level are often intra-class within broader categories, and thus do not significantly reflect meaningful failures neither impact performance under coarse evaluation. However, the magnitude of the improvement varies across architectures, suggesting that coarse-label evaluation captures complementary information beyond fine-grained segmentation accuracy and provides additional insight into the semantic consistency of model predictions.

On nuScenes~\cite{caesar2020nuscenes}, point-based methods achieve the strongest coarse-label performance. PTv3~\cite{wu2024ptv3} attains 87.8\% mIoU$^{CO}_{LA}$ and leads across most semantic groups, including \textit{vehicle}, \textit{living-beings}, \textit{driveable-ground}, and \textit{structure}. Among projection-based approaches, FRNet~\cite{xu2025frnet} remains highly competitive with 86.3\%, achieving the best performance on \textit{nature} and strong results on \textit{structure} and \textit{other-ground}. HARP-NeXt reaches 86.7\% while offering substantially higher throughput (100 FPS).

\begin{table*}[h]
    \centering
    \caption{Models' coarse labels evaluation on nuScenes \cite{caesar2020nuscenes}. The \textbf{best} and \underline{second best} scores are in \textbf{bold} and \underline{underline}. FPS measurements are reported on a single NVIDIA GeForce RTX 4090 GPU.}
    \label{tab:coarseperformancenuscenes}
    \resizebox{0.85\textwidth}{!}{
    \begin{tabular}{r| c | c| c | c | c c c c c c c}
        \toprule
        \textbf{Methods} & \textbf{Category} & \textbf{FPS} & \rotatebox{70}{\textbf{mIoU$^{FG}$}} & \rotatebox{70}{\textbf{mIoU$^{CO}_{LA}$}} & 
        \rotatebox{70}{\textbf{vehicle}} &
        \rotatebox{70}{\textbf{living-be.}} &
        \rotatebox{70}{\textbf{driv-gr.}} &
        \rotatebox{70}{\textbf{other-gr.}} &
        \rotatebox{70}{\textbf{structure}} &
        \rotatebox{70}{\textbf{nature}} &
        \rotatebox{70}{\textbf{static-obj.}} \\
        
        \midrule
        \midrule
        WaffleIron~\cite{puy2023waffleiron} & \multirow{2}{*}{Point-based} & 9.0 & 76.1 & 82.4 & 89.9 & 75.8 & 96.7 & 75.7 & 87.5 & 84.7 & 66.2 \\ 
        PTv3~\cite{wu2024ptv3} &  & 4.1 & \textbf{78.4} & \textbf{87.8} & \textbf{96.1} & \textbf{82.8} & \textbf{97.8} & 77.7 & \textbf{93.5} & \underline{89.0} & \textbf{77.7} \\
        \midrule
        SalsaNext~\cite{cortinhal2020salsanext} & \multirow{3}{*}{Projection-based}  & \underline{76.9} & 68.2 & 78.1 & 88.2 & 71.1 & 95.6 & 72.5 & 83.2 & 79.4 & 57.1 \\
        CENet~\cite{cheng2022cenet} &  & 62.5 & 73.3 & 79.4 & 90.1 & 68.0 & 95.4 & 73.9 & 85.9 & 83.8 & 58.8 \\
        FRNet~\cite{xu2025frnet} & & 12.2 & 75.1 & 86.3 & \underline{94.8} & 75.7 & \underline{97.6} & \underline{80.3} & \underline{92.7} & \textbf{91.0} & 72.2 \\
        \midrule
        Minkowski~\cite{choy2019minkowski} & Sparse Conv-based & 21.3 & 73.5 & 83.8 & 91.6 & 76.3 & 96.5 & 76.5 & 88.9 & 87.7 & 68.8 \\
        \midrule
        SPVCNN~\cite{tang2020spvnas} & \multirow{3}{*}{Fusion-based} & 17.5 & 72.6 & 82.7 & 90.6 & 75.5 & 96.1 & 73.6 & 88.4 & 86.2 & 68.3 \\
        HARP-NeXt~\cite{abouhaidar2025harpnext} &  & \textbf{100} & \underline{77.1} & \underline{86.7} & 94.0 & \underline{78.3} & \underline{97.6} & \textbf{80.5} & 92.6 & \textbf{91.0} & \underline{73.2} \\
        \bottomrule
    \end{tabular}}
\end{table*}

\begin{table*}[h]
    \centering
    \caption{Models' coarse labels evaluation on SemanticKITTI \cite{behley2019semantickitti}. The \textbf{best} and \underline{second best} scores are in \textbf{bold} and \underline{underline}. FPS measurements are reported on a single NVIDIA GeForce RTX 4090 GPU.}
    \label{tab:coarseperformancesemantickitti}
    \resizebox{0.85\textwidth}{!}{
    \begin{tabular}{r| c | c| c | c | c c c c c c c}
        \toprule
        \textbf{Methods} & \textbf{Category} & \textbf{FPS} & \rotatebox{70}{\textbf{mIoU$^{FG}$}} & \rotatebox{70}{\textbf{mIoU$^{CO}_{LA}$}} & 
        \rotatebox{70}{\textbf{vehicle}} &
        \rotatebox{70}{\textbf{living-be.}} &
        \rotatebox{70}{\textbf{driv-gr.}} &
        \rotatebox{70}{\textbf{other-gr.}} &
        \rotatebox{70}{\textbf{structure}} &
        \rotatebox{70}{\textbf{nature}} &
        \rotatebox{70}{\textbf{static-obj.}} \\
        
        \midrule
        \midrule
        WaffleIron~\cite{puy2023waffleiron} & Point-based & 2.7 & \underline{65.8} & \textbf{87.4} & \textbf{97.1} & \underline{83.2} & 92.8 & 81.9 & \underline{91.2} & \underline{94.2} & 71.6\\ 
        \midrule
        SalsaNext~\cite{cortinhal2020salsanext} & \multirow{3}{*}{Projection-based} & \underline{27.8} & 55.9 & 74.4 & 86.4 & 48.1 & 87.6 & 70.6 & 78.7 & 88.2 & 61.8 \\
        CENet~\cite{cheng2022cenet} &  & 17.2 & 62.6 & 83.9 & 91.2 & 76.5 & 91.8 & 79.7 & 86.9 & 92.8 & 69.0\\
        FRNet~\cite{xu2025frnet} & & 11.6 & \textbf{66.0} & 83.3 & \textbf{97.1} & 55.6 & \underline{93.2} & \underline{82.3} & 90.4 & 93.8 & 70.6\\
        \midrule
        Minkowski~\cite{choy2019minkowski} & Sparse Conv-based & 14.1 & 64.3 & \underline{87.3} & \textbf{97.1} & \textbf{83.5} & 91.3 & 80.4 & \underline{91.2} & \underline{94.2} & \underline{73.5} \\
        \midrule
        SPVCNN~\cite{tang2020spvnas} & \multirow{2}{*}{Fusion-based} & 11.8 & 65.3 & 86.8 & \underline{96.8} & \underline{83.2} & 91.0 & 79.6 & 90.6 & 93.8 & 72.4\\
        HARP-NeXt~\cite{abouhaidar2025harpnext} &  & \textbf{76.9} & 65.1 & 86.2 & 96.5 & 64.2 & \textbf{96.5} & \textbf{86.1} & \textbf{91.9} & \textbf{94.3} & \textbf{73.8} \\
        \bottomrule
    \end{tabular}}
\end{table*}

These differences in accuracy across state-of-the-art methods are accompanied by significant variations in computational cost. Point-based methods suffer from low runtime speed (e.g., PTv3 at 4.1 FPS), limiting their applicability in real-time applications. In contrast, projection-based approaches offer substantially higher throughput, with SalsaNext achieving 76.9 FPS at the expense of reduced accuracy. Fusion-based methods, such as HARP-NeXt, provide a better trade-off, balancing performance and efficiency.

On SemanticKITTI~\cite{behley2019semantickitti}, similar trends are observed. WaffleIron~\cite{puy2023waffleiron} achieves the highest coarse-level performance (87.4\% mIoU$^{CO}_{LA}$), followed closely by Minkowski (87.3\%) and SPVCNN (86.8\%), indicating strong generalization under coarse labels. Projection-based methods show greater variability: although FRNet~\cite{xu2025frnet} achieves the highest fine-grained mIoU$^{FG}$ (66.0\%), its coarse performance (83.3\%) is comparatively lower. Fusion-based approaches remain competitive, with HARP-NeXt reaching 86.2\% while providing the highest efficiency (76.9 FPS) and strong performance on key classes such as \textit{driveable ground} and \textit{structure}.

A notable observation on SemanticKITTI~\cite{behley2019semantickitti} is that FRNet~\cite{xu2025frnet}, despite achieving the highest fine-grained mIoU$^{FG}$ (66.0\%), attains a comparatively lower coarse-label performance (83.3\%). This behavior is largely driven by its \textit{living-being} score (55.6\%), where the projection-based representation struggles with the sparse and small-scale point clouds of pedestrians and cyclists. In contrast, voxel-based methods such as Minkowski~\cite{choy2019minkowski} and SPVCNN~\cite{tang2020spvnas} achieve substantially higher performance on this category (83.5\% and 83.2\%, respectively), suggesting a stronger ability to capture the geometric characteristics of vulnerable road users. Importantly, the \textit{living-being} category is among the most safety-critical semantic groups in autonomous systems. Consequently, strong overall segmentation accuracy does not necessarily translate into superior suitability for real-world deployment if performance degrades on classes that are critical for safe navigation and collision avoidance.

Overall, the COLA benchmark highlights an important aspect of robustness: modern LiDAR segmentation models are generally resilient to changes in label granularity. This is desirable for real-world deployment, where annotation schemes may vary across applications. Nevertheless, robustness must be considered jointly with computational efficiency and performance on safety-critical categories. Projection-based methods consistently achieve lower performance on the \textit{living-being} category, whereas point-based methods provide stronger category-level semantic understanding at a significantly higher computational cost. Fusion-based approaches offer a more balanced trade-off between accuracy, safety-critical perception, and real-time performance.

\subsection{Robustness evaluation against adverse conditions}
\label{subsec:robustnessresults}

The quantitative robustness results on nuScenes-C and SemanticKITTI-C are reported in Table~\ref{tab:robustness}, with corruptions grouped into three degradation categories: atmospheric (fog, wet ground, snow), geometric (motion blur, beam missing), and sensor (LiDAR crosstalk, incomplete echo, cross-sensor).

On nuScenes-C, most methods exhibit moderate degradation across corruption categories, revealing distinct robustness characteristics across architectural families. Sparse convolution-based methods achieve the strongest robustness to geometric corruptions, with MinkowskiNet~\cite{choy2019minkowski} obtaining the highest geometric group average. This is due to the spatial aggregation performed within voxelized representations, which reduces sensitivity to structural perturbations such as motion blur and missing beam. Projection-based methods show more heterogeneous performance. Although both FRNet~\cite{xu2025frnet} and CENet~\cite{cheng2022cenet} operate on range-image projections, FRNet achieves substantially better robustness, obtaining the lowest mCE among non-baseline methods (103.7\%) and stronger atmospheric and sensor corruption averages. This suggests that retaining and exploiting fine-grained point-level features in addition to the projected representation in FRNet's backbone improves robustness by preserving geometric cues that may be degraded or lost during projection. Fusion-based methods generally provide a favorable balance between accuracy, robustness, and efficiency. In particular, HARP-NeXt~\cite{abouhaidar2025harpnext} achieves the highest overall clean mIoU (77.1\%), the best atmospheric corruption group average (71.2\%), and competitive robustness (mCE = 105.7\%) across all corruption groups while operating at 14.1 FPS on the Jetson AGX Orin, substantially faster than other top-performing methods. This suggests that lightweight fusion mechanisms can retain many of the robustness benefits of multi-representation learning without incurring the computational cost typically associated with more complex architectures. More broadly, the robustness--efficiency trade-off varies considerably across methods, indicating that robustness should be evaluated jointly with computational requirements when considering deployment in real-world autonomous driving systems. Finally, across all architectural categories, sensor-related corruptions remain particularly challenging, highlighting the limited robustness of current methods to sensor-specific artifacts and cross-sensor domain shifts.

\begin{table*}[htbp]
    \centering
    \caption{Robustness evaluation on the nuScenes-C and SemanticKITTI-C benchmarks~\cite{kong2023robo3d}. mCE and mIoU are reported in (\%), while FPS denotes inference throughput measured on a NVIDIA Jetson AGX Orin. Group averages correspond to the mean mIoU across corruption severities. \textbf{Bold} and \underline{underline} indicate the best and second-best results, respectively.}
    \label{tab:robustness}
    \resizebox{0.99\textwidth}{!}{
    \begin{tabular}{c|r|c|c|c|c|ccc|cc|ccc|ccc}
        \toprule
         & \multirow{2}{*}{\textbf{Methods}} & \multirow{2}{*}{\textbf{Category}} & \multirow{2}{*}{\textbf{FPS}} & \multirow{2}{*}{\textbf{mCE$\downarrow$}} & \multirow{2}{*}{\textbf{mIoU$\uparrow$}} & \multicolumn{3}{c|}{\textbf{Atmospheric}} & \multicolumn{2}{c|}{\textbf{Geometric}} & \multicolumn{3}{c|}{\textbf{Sensor}} & \multicolumn{3}{c}{\textbf{Group Averages}} \\
         & & & & & & \textbf{Fog} & \textbf{Wet} & \textbf{Snow} & \textbf{Mot.} & \textbf{Beam} & \textbf{Cross.} & \textbf{Echo} & \textbf{Sens.} & \textbf{Atmos.} & \textbf{Geom.} & \textbf{Sens.} \\
        \midrule
        \midrule
        \multirow{7}{*}{\rotatebox{90}{\textbf{nuScenes-C}}}
        & WaffleIron~\cite{puy2023waffleiron} & Point-based & 1.9 & 106.7 & \underline{76.1} & 56.1 & 73.9 & 49.6 & \underline{59.5} & 65.2 & 33.1 & 61.5 & 44.0 & 59.9 & \underline{62.4} & 46.2 \\
        \cmidrule{2-17}
        & CENet~\cite{cheng2022cenet} & \multirow{2}{*}{Proj-based} & 10.3 & 112.8 & 73.3 & 67.0 & 69.9 & 61.6 & 58.3 & 50.0 & 60.9 & 53.3 & 24.8 & 66.2 & 54.2 & 46.3 \\
        & FRNet~\cite{xu2025frnet} &  & 2.6 & \underline{103.7} & 75.1 & \underline{71.2} & \underline{74.1} & \textbf{66.5} & 39.9 & 67.7 & \underline{65.8} & 57.5 & 41.3 & \underline{70.6} & 53.8 & \textbf{54.9} \\
        \cmidrule{2-17}
        & Minkowski~\cite{choy2019minkowski} & \multirow{2}{*}{Sparse-based} & 6.8 & \textbf{100.0} & 73.5 & 53.6 & 73.9 & 40.4 & \textbf{73.4} & \textbf{68.5} & 26.6 & \textbf{63.8} & \textbf{51.0} & 55.9 & \textbf{70.9} & 47.1 \\
        & Cylinder3D~\cite{zhu2021cylindrical} &  & --- & 111.8 & 76.1 & 59.9 & 72.7 & 58.1 & 42.1 & 64.5 & 44.4 & 60.5 & 42.2 & 63.6 & 53.3 & 49.0 \\
        \cmidrule{2-17}
        & SPVCNN~\cite{tang2020spvnas} & \multirow{2}{*}{Fusion-based} & 5.9 & 106.7 & 72.6 & 59.0 & 72.5 & 41.1 & 58.4 & 65.4 & 36.8 & 62.3 & \underline{49.2} & 57.5 & 61.9 & 49.4 \\
        & HARP-NeXt~\cite{abouhaidar2025harpnext} &  & \textbf{14.1} & 105.7 & \textbf{77.1} & \textbf{72.2} & \textbf{75.9} & \underline{65.6} & 34.9 & \underline{67.8} & \textbf{67.3} & 56.1 & 39.6 & \textbf{71.2} & 51.4 & \underline{54.3} \\
        \midrule
        \midrule
        \multirow{9}{*}{\rotatebox{90}{\textbf{SemanticKITTI-C}}}
        & KPConv~\cite{thomas2019kpconv} & \multirow{2}{*}{Point-based} & --- & \textbf{99.5} & 62.2 & 54.5 & 57.7 & \textbf{54.1} & 25.7 & 57.3 & 53.4 & 55.6 & \underline{53.9} & \textbf{55.4} & 41.5 & \underline{54.3} \\
        & WaffleIron~\cite{puy2023waffleiron} & & 0.5 & 109.5 & \underline{65.8} & 45.5 & 58.6 & 49.3 & 33.0 & \textbf{59.3} & 22.5 & \textbf{58.6} & \textbf{54.6} & 51.1 & 46.2 & \textbf{55.9} \\
        \cmidrule{2-17}
        & SalsaNext~\cite{cortinhal2020salsanext} & \multirow{3}{*}{Proj-based} & \textbf{9.2} & 116.1 & 55.9 & 34.9 & 48.4 & 45.6 & 47.9 & 48.6 & 40.2 & 48.0 & 44.7 & 42.9 & 48.3 & 44.3 \\
        & CENet~\cite{cheng2022cenet} & & 6.1 & 103.4 & 62.6 & 42.7 & 57.3 & \underline{53.6} & 52.7 & 55.8 & 45.4 & 53.4 & 45.8 & 51.2 & 54.3 & 48.2 \\
        & FRNet~\cite{xu2025frnet} &  & 2.5 & \textbf{99.5} & \textbf{66.0} & 50.6 & \textbf{59.2} & 49.5 & \textbf{54.9} & 57.3 & 43.9 & \underline{55.8} & 51.1 & 53.1 & \textbf{56.1} & 50.3 \\
        \cmidrule{2-17}
        & Minkowski~\cite{choy2019minkowski} & \multirow{2}{*}{Sparse-based} & 4.7 & \underline{100.0} & 64.3 & \textbf{55.9} & 54.0 & 53.3 & 32.9 & 56.3 & \textbf{58.3} & 54.4 & 46.1 & \underline{54.4} & 44.6 & 52.9 \\
        & Cylinder3D~\cite{zhu2021cylindrical} &  & --- & 103.3 & 63.2 & 37.1 & 57.5 & 46.9 & 52.5 & \underline{57.6} & 56.0 & 52.5 & 46.2 & 47.2 & \underline{55.1} & 51.6 \\
        \cmidrule{2-17}
        & SPVCNN~\cite{tang2020spvnas} & \multirow{2}{*}{Fusion-based} & 4.0 & 100.3 & 65.3 & \underline{55.3} & 54.0 & 51.4 & 34.5 & 56.7 & \underline{58.1} & 54.6 & 46.0 & 53.6 & 45.6 & 52.9 \\
        & HARP-NeXt~\cite{abouhaidar2025harpnext} &  & \underline{8.3} & 105.9 & 65.1 & 45.8 & \underline{58.8} & 46.0 & 43.2 & 57.0 & 43.8 & 49.9 & 50.6 & 50.2 & 50.1 & 48.1 \\
        \bottomrule
    \end{tabular}}
\end{table*}

On SemanticKITTI-C, the robustness gap between methods becomes more pronounced. Point-based methods demonstrate particularly strong robustness, with KPConv~\cite{thomas2019kpconv} achieving the lowest mCE (99.5\%) and the highest atmospheric corruption group average (55.4\%), suggesting that local neighborhood aggregation is especially resilient to perturbations affecting point density and geometric structure. Projection-based methods exhibit more variable behavior. FRNet~\cite{xu2025frnet} achieves the highest clean mIoU (66.0\%), the best geometric group average (56.1\%), and the lowest mCE among non-baseline methods, indicating that preserving point-level geometric information alongside the projected representation improves robustness under severe corruptions. However, this comes at a substantial computational cost, with FRNet operating at only 2.5 FPS on the Jetson AGX Orin. In contrast, SalsaNext~\cite{cortinhal2020salsanext} achieves the highest throughput (9.2 FPS) among the evaluated methods but experiences substantially larger degradation across all corruption categories. Sparse convolution-based methods remain competitive, achieving balanced robustness without exhibiting a strong performance for any particular corruption group, while fusion-based approaches provide a balanced compromise between robustness and computational efficiency. Compared to its performance on nuScenes-C, HARP-NeXt~\cite{abouhaidar2025harpnext} maintains high efficiency (8.3 FPS) but exhibits a larger robustness degradation on SemanticKITTI-C, suggesting that fusion-based representations remain sensitive to perturbations affecting fine-grained geometric structure.

The variation in model robustness across nuScenes-C and SemanticKITTI-C can be attributed to dataset characteristics. SemanticKITTI~\cite{behley2019semantickitti} contains denser point clouds acquired with a Velodyne HDL-64E sensor, which encourage learning of fine-grained geometric structures. As a result, corruptions that affect geometric completeness or introduce structural inconsistencies have a stronger impact across semantic segmentation methods. In contrast, the sparser scans in nuScenes~\cite{caesar2020nuscenes} (Velodyne HDL-32E) reduce reliance on fine geometric details, leading to comparatively more stable results under certain corruption types. Overall, models exhibit increased sensitivity to geometric and sensor degradations on SemanticKITTI-C compared to nuScenes-C.

The evaluation under adverse conditions indicates that robustness remains a key challenge for 3D semantic segmentation. Many state-of-the-art methods overfit to clean data distributions, limiting their generalization to real-world noisy scenarios. Improving robustness requires both architectural advances and more effective training strategies, such as incorporating realistic corruption-aware data augmentation (e.g., motion blur simulation, beam dropping, and non-uniform sub-sampling) to better bridge the gap between clean benchmarks and real-world conditions. Importantly, the results also highlight a clear robustness--efficiency trade-off, where higher robustness is often achieved at the cost of reduced inference speed, making real-time deployment a key constraint in practical autonomous driving systems.

\subsection{Domain generalization evaluation}
\label{subsec:domain_generalization_results}

\begin{table*}[htbp]
\caption{Domain generalization results from nuScenes \cite{caesar2020nuscenes} (source) to ParisLuco3D \cite{sanchez2024parisluco3d} (target).}
\label{tab:domaingeneralization_NS_PL}
\centering
\resizebox{\textwidth}{!}{
\begin{tabular}{c|c*{16}{|c}} 
\toprule
\multicolumn{1}{c|}{\textbf{Method}} & \multicolumn{1}{c|}{\textbf{mIoU$^{PL}$}} & 
\rotatebox{70}{\textbf{barrier}} & 
\rotatebox{70}{\textbf{bicycle}} & 
\rotatebox{70}{\textbf{bus}} & 
\rotatebox{70}{\textbf{car}} & 
\rotatebox{70}{\textbf{const. veh.}} & 
\rotatebox{70}{\textbf{motorcycle}} & 
\rotatebox{70}{\textbf{pedestrian}} & 
\rotatebox{70}{\textbf{traffic cone}} & 
\rotatebox{70}{\textbf{trailer}} & 
\rotatebox{70}{\textbf{truck}} & 
\rotatebox{70}{\textbf{driv. surface}} & 
\rotatebox{70}{\textbf{other flat}} & 
\rotatebox{70}{\textbf{sidewalk}} & 
\rotatebox{70}{\textbf{terrain}} & 
\rotatebox{70}{\textbf{manmade}} & 
\rotatebox{70}{\textbf{vegetation}} \\
\toprule

KPConv~\cite{thomas2019kpconv} & 11.4 & \textbf{4.2} & 0.1 & 2.1 & 8.3 & 0.2 & 1.9 & 11.4 & 0.0 & 0.0 & 1.7 & 8.3 & 0.0 & 1.7 & 3.7 & 73.2 & 65.7 \\

CENet~\cite{cheng2022cenet} & 31.6 & \underline{4.1} & \textbf{4.7} & 35.7 & 61.6 & 1.6 & \underline{22.7} & 51.6 & 0.0 & 0.0 & 6.1 & \underline{77.4} & \textbf{22.5} & \underline{56.7} & \underline{13.3} & 81.1 & 66.6 \\

FRNet~\cite{xu2025frnet} & 28.5 & 3.2 & \underline{1.2} & 34.1 & \underline{66.5} & 0.1 & 20.1 & 48.4 & 0.6 & 0.0 & 9.2 & 69.7 & \underline{18.1} & 55.4 & 10.9 & 73.4 & 44.5 \\

Minkowski~\cite{choy2019minkowski}  & \textbf{32.5} & 1.8 & 0.4 & \underline{48.4} & \textbf{78.1} & \underline{3.9} & 10.6 & \underline{51.7} & 0.3 & 0.0 & \textbf{20.4} & 72.7 & 2.5 & 47.4 & \textbf{14.3} & \underline{83.5} & \textbf{83.2} \\

Cylinder3D~\cite{zhu2021cylindrical} & 17.1 & 0.3 & 0.0 & 5.0 & 31.5 & 0.4 & 0.1 & 17.4 & \underline{1.2} & 0.0 & 14.0 & 13.3 & 0.1 & 25.5 & 5.8 & 77.3 & \underline{82.4} \\

SPVCNN~\cite{tang2020spvnas} & \underline{32.1} & 1.7 & 0.8 & \textbf{49.5} & 66.0 & \textbf{5.4} & \textbf{27.1} & \textbf{55.9} & 0.3 & 0.0 & \underline{15.4} & 69.3 & 1.4 & 42.8 & 11.8 & \textbf{84.8} & 82.1 \\

HARP-NeXt~\cite{abouhaidar2025harpnext}  & 28.7 & 2.7 & 1.1 & 32.5 & 64.4 & 0.1 & 19.0 & 47.5 & \textbf{1.7} & 0.0 & 12.7 & \textbf{77.6} & 6.9 & \textbf{57.1} & 11.4 & 74.5 & 50.9 \\

\bottomrule
\end{tabular}}
\end{table*}

\begin{table*}[htbp]
\caption{Domain generalization results from SemanticKITTI \cite{behley2019semantickitti} (source) to ParisLuco3D \cite{sanchez2024parisluco3d} (target).}
\label{tab:domaingeneralization_SK_PL}
\centering
\resizebox{\textwidth}{!}{
\begin{tabular}{c|c*{17}{|c}} 
\toprule

\multicolumn{1}{c|}{\textbf{Method}} & \multicolumn{1}{c|}{\textbf{mIoU$^{PL}$}} & 
\rotatebox{70}{\textbf{car}} & 
\rotatebox{70}{\textbf{bicycle}} & 
\rotatebox{70}{\textbf{motorcycle}} & 
\rotatebox{70}{\textbf{truck}} & 
\rotatebox{70}{\textbf{other-vehicle}} & 
\rotatebox{70}{\textbf{person}} & 
% \rotatebox{70}{\textbf{bicyclist}} & 
% \rotatebox{70}{\textbf{motorcyclist}} & 
\rotatebox{70}{\textbf{road}} & 
\rotatebox{70}{\textbf{parking}} & 
\rotatebox{70}{\textbf{sidewalk}} & 
\rotatebox{70}{\textbf{other-ground}} & 
\rotatebox{70}{\textbf{building}} & 
\rotatebox{70}{\textbf{fence}} & 
\rotatebox{70}{\textbf{vegetation}} & 
\rotatebox{70}{\textbf{trunk}} & 
\rotatebox{70}{\textbf{terrain}} & 
\rotatebox{70}{\textbf{pole}} & 
\rotatebox{70}{\textbf{traffic-sign}} \\
\toprule

KPConv~\cite{thomas2019kpconv} & 22.1 & 39.8 & \underline{7.4} & 9.1 & 0.3 & 5.1 & \textbf{30.6} & 8.1 & 0.1 & 41.5 & 0.7 & 58.5 & 11.7 & 66.9 & \textbf{49.3} & \underline{14.0} & 25.6 & 6.8 \\

CENet~\cite{cheng2022cenet} & 2.9 & 2.2 & 0.0 & 0.0 & 0.4 & 0.0 & 0.0 & 1.6 & 0.0 & 6.8 & 0.0 & 10.1 & 13.1 & 10.6 & 0.0 & 1.1 & 2.6 & 1.3 \\

FRNet~\cite{xu2025frnet} & 3.4 & 1.3 & 0.0 & 0.0 & 0.0 & 0.0 & 0.0 & 0.1 & 0.0 & 21.3 & 0.0 & 31.5 & 1.5 & 1.1 & 0.0 & 0.7 & 0.2 & 0.0 \\

Minkowski~\cite{choy2019minkowski} & \textbf{33.3} & \textbf{69.5} & \textbf{9.3} & \textbf{15.4} & \textbf{4.1} & \textbf{32.1} & 20.5 & \underline{71.4} & \underline{1.0} & \underline{66.7} & 1.5 & \underline{67.3} & 18.1 & \underline{71.4} & \underline{44.2} & \textbf{15.7} & \textbf{40.2} & \underline{19.0} \\

Cylinder3D~\cite{zhu2021cylindrical} & 25.5 & 46.4 & 4.6 & 5.8 & 0.3 & 15.2 & 11.3 & 58.9 & \textbf{3.9} & 57.2 & \underline{1.7} & 65.8 & \textbf{36.6} & 54.5 & 24.4 & 10.8 & 31.7 & 3.8 \\

SPVCNN~\cite{tang2020spvnas} & \underline{31.5} & \underline{66.7} & 7.0 & \underline{14.0} & \underline{4.0} & \underline{18.9} & \underline{21.8} & 66.6 & 0.2 & \textbf{67.0} & 0.1 & 66.3 & 13.0 & \textbf{71.6} & 43.2 & 10.8 & \underline{38.3} & \textbf{25.4} \\

HARP-NeXt~\cite{abouhaidar2025harpnext} & 18.8 & 38.4 & 0.5 & 0.1 & 0.0 & 4.0 & 1.2 & \textbf{79.5} & 0.7 & 48.0 & \textbf{1.9} & \textbf{70.4} & \underline{34.1} & 29.3 & 3.3 & 1.5 & 3.8 & 3.1 \\

\bottomrule
\end{tabular}}

\end{table*}

Complementing the corruption-based robustness analysis, we evaluate cross-domain generalization under real-world distribution shifts using the ParisLuco3D benchmark~\cite{sanchez2024parisluco3d}. We follow a single-source to single-target protocol, where models are trained on a source dataset and directly evaluated on the unseen ParisLuco3D (${PL}$) benchmark without any adaptation or access to target-domain data. ParisLuco3D is designed for cross-domain generalization and enables consistent performance comparison across source datasets. Its fine-grained label taxonomy can be mapped to both nuScenes~\cite{caesar2020nuscenes} and SemanticKITTI~\cite{behley2019semantickitti}, which we use as source-domain training datasets. Accordingly, we use dataset-specific fine-grained label mappings from ParisLuco3D~\cite{sanchez2024parisluco3d}, and consider two complementary domain shifts:

\textbf{(i) nuScenes (source) $\rightarrow$ ParisLuco3D (target)}

We first evaluate domain generalization from nuScenes~\cite{caesar2020nuscenes} to ParisLuco3D~\cite{sanchez2024parisluco3d}, as reported in \cref{tab:domaingeneralization_NS_PL}. Both datasets use the same Velodyne HDL-32E LiDAR sensor; therefore, LiDAR intensity is retained for all models. This configuration enables \textit{appearance} and \textit{scene shifts} without introducing \textit{sensor shift}. These include differences in urban layout, geographic context, and environmental conditions. Notably, ParisLuco3D contains sequences recorded in adverse weather, including rain, leading to distorted point cloud structures (e.g., pedestrians with umbrellas) and partial road occlusions due to water accumulation, further increasing the challenge of robust perception.

\Cref{tab:domaingeneralization_NS_PL} shows that this domain shift remains challenging, as no method achieves consistently high performance across classes. Despite identical sensing conditions, overall performance is limited, with Minkowski~\cite{choy2019minkowski} and SPVCNN~\cite{tang2020spvnas} obtaining the best mIoU scores of 32.5\% and 32.1\%, respectively. This indicates that appearance and scene variations alone can significantly degrade performance, even when point cloud acquisition settings are identical.

Several methods underperform substantially. KPConv~\cite{thomas2019kpconv} reaches 11.4\% mIoU, while Cylinder3D~\cite{zhu2021cylindrical} achieves 17.1\%, highlighting the difficulty of generalizing to weather-affected scenes in ParisLuco3D. Projection-based methods, including CENet~\cite{cheng2022cenet} and FRNet~\cite{xu2025frnet}, achieve competitive results on dominant classes but are less stable overall due to their sensitivity to range-image representations under environmental shifts. Fusion-based approaches such as HARP-NeXt~\cite{abouhaidar2025harpnext}, which also incorporates a range-image projection, show similar behavior, indicating that reliance on range-view features remains limiting under appearance shifts. Sparse convolution-based methods, particularly Minkowski~\cite{choy2019minkowski} and SPVCNN~\cite{tang2020spvnas}, perform best overall, suggesting that geometry-preserving voxel representations offer improved robustness under domain shift, though performance remains uneven across classes.

At the class level, large and structured categories such as \textit{driveable surface}, \textit{sidewalk}, \textit{manmade}, and \textit{vegetation} consistently achieve higher IoU scores due to stable geometry and spatial extent. In contrast, small or rare classes such as \textit{bicycle}, \textit{traffic cone}, and \textit{trailer} remain poorly segmented across methods, reflecting challenges from class imbalance, weak geometric cues, and occlusions.

\textbf{(ii) SemanticKITTI (source) $\rightarrow$ ParisLuco3D (target)}

We then consider domain shift from SemanticKITTI~\cite{behley2019semantickitti} to ParisLuco3D~\cite{sanchez2024parisluco3d}. This also introduces a significant \textit{sensor shift} on top of the \textit{scene shift} and \textit{appearance shift}, as SemanticKITTI is acquired using a Velodyne HDL-64E LiDAR with higher vertical resolution and a different field-of-view than the Velodyne HDL-32E. This affects point density and sampling patterns, leading to additional distortions in cross-dataset generalization. As a result, this setup provides a more challenging evaluation of domain generalization, requiring models to perform under simultaneous variations in sensor configuration, environment, and scene composition.

\Cref{tab:domaingeneralization_SK_PL} highlights that generalization from SemanticKITTI \cite{behley2019semantickitti} to ParisLuco3D \cite{sanchez2024parisluco3d} remains a considerable challenge. Sparse convolution-based methods again achieve the best results, with Minkowski \cite{choy2019minkowski} and SPVCNN \cite{tang2020spvnas} leading at 33.3\% and 31.5\% mIoU, respectively. The fusion-based method HARP-NeXt attains 18.8\%, which is modest overall; however, compared to the projection-based methods CENet \cite{cheng2022cenet} and FRNet \cite{xu2025frnet}, which yield near-zero performance, HARP-NeXt demonstrates better robustness, as a method also incorporating a range-image projection within its architecture. This suggests that, unlike ResNet-based backbones in CENet and FRNet that tightly couple spatial-channel feature learning on range images and tend to overfit source-domain patterns, HARP-NeXt’s depth-wise separable convolutions promote more channel-specific representations, improving generalization under domain shift.

Overall, most methods exhibit stronger generalization from nuScenes~\cite{caesar2020nuscenes} to ParisLuco3D~\cite{sanchez2024parisluco3d} than from SemanticKITTI~\cite{behley2019semantickitti}, largely due to the identical sensor configuration in the former setup. However, KPConv~\cite{thomas2019kpconv} and Cylinder3D~\cite{zhu2021cylindrical} deviate from this trend, achieving higher mIoU when trained on SemanticKITTI~\cite{behley2019semantickitti}. This is attributed to their reliance on local geometric structure rather than sensor-specific patterns, where KPConv learns continuous point-based kernels and Cylinder3D encodes occupancy in cylindrical space. Both benefit from the richer geometric supervision of SemanticKITTI, leading to more transferable representations. These results indicate that generalization is influenced not only by sensor and scene similarity but also by architectural designs. Finally, regarding domain generalization, current approaches remain insufficient for reliable real-world deployment, underscoring the need for sensor-invariant architectures that can generalize across diverse LiDAR configurations, environments, and scene shifts.

\section{CONCLUSION}
\label{sec:conclusion}

In this work, we presented a unified evaluation protocol for assessing the deployment readiness of LiDAR semantic segmentation models through three complementary dimensions: coarse-label safety assessment, robustness to real-world corruptions, and cross-domain generalization, alongside inference runtime measurements on an embedded Jetson AGX Orin and an RTX 4090 GPU.

Our results show that clean-benchmark accuracy is a poor predictor of deployment readiness. Methods that rank highly on standard metrics can fail on safety-critical classes such as \textit{living-being}, degrade substantially under sensor corruptions, or suffer significant performance drops under domain shifts, with projection-based methods generally exhibiting greater sensitivity to sensor and scene variations. No single state-of-the-art method achieves a strong trade-off between accuracy, robustness, generalization, and computational efficiency.

Our proposed protocol provides a reproducible framework for evaluating LiDAR semantic segmentation models beyond conventional benchmarks, enabling a more comprehensive assessment of deployment readiness, promoting more realistic evaluation practices, and guiding future model design for real-world autonomous driving perception.

\addtolength{\textheight}{-12cm}   % This command serves to balance the column lengths
                                  % on the last page of the document manually. It shortens
                                  % the textheight of the last page by a suitable amount.
                                  % This command does not take effect until the next page
                                  % so it should come on the page before the last. Make
                                  % sure that you do not shorten the textheight too much.

\bibliographystyle{IEEEtran} % IEEEtran already built in no need to upload
\bibliography{iros_main}

@INPROCEEDINGS{puy2023waffleiron,
  author={Puy, Gilles and Boulch, Alexandre and Marlet, Renaud},
  booktitle={IEEE/CVF International Conference on Computer Vision (ICCV)}, 
  title={Using a Waffle Iron for Automotive Point Cloud Semantic Segmentation}, 
  year={2023}
}

@INPROCEEDINGS{thomas2019kpconv,
  author={Thomas, Hugues and Qi, Charles R. and Deschaud, Jean-Emmanuel and Marcotegui, Beatriz and Goulette, François and Guibas, Leonidas},
  booktitle={IEEE/CVF International Conference on Computer Vision (ICCV)}, 
  title={KPConv: Flexible and Deformable Convolution for Point Clouds}, 
  year={2019}
}

@inproceedings{tang2020spvnas,
author="Tang, Haotian
and Liu, Zhijian
and Zhao, Shengyu
and Lin, Yujun
and Lin, Ji
and Wang, Hanrui
and Han, Song",
title="Searching Efficient 3D Architectures with Sparse Point-Voxel Convolution",
booktitle="European Conference on Computer Vision (ECCV)",
year="2020"
}

@inproceedings{choy2019minkowski,
author = {Choy, Christopher and Gwak, JunYoung and Savarese, Silvio},
title = {4D Spatio-Temporal ConvNets: Minkowski Convolutional Neural Networks},
booktitle = {IEEE/CVF Conference on Computer Vision and Pattern Recognition (CVPR)},
year = {2019}
}

@INPROCEEDINGS{zhu2021cylindrical,
  author={Zhu, Xinge and Zhou, Hui and Wang, Tai and Hong, Fangzhou and Ma, Yuexin and Li, Wei and Li, Hongsheng and Lin, Dahua},
  booktitle={IEEE/CVF Conference on Computer Vision and Pattern Recognition (CVPR)}, 
  title={Cylindrical and Asymmetrical 3D Convolution Networks for LiDAR Segmentation}, 
  year={2021},
}

@inproceedings{behley2019semantickitti,
  author={Behley, Jens and Garbade, Martin and Milioto, Andres and Quenzel, Jan and Behnke, Sven and Stachniss, Cyrill and Gall, Jürgen},
  booktitle={IEEE/CVF International Conference on Computer Vision (ICCV)}, 
  title={SemanticKITTI: A Dataset for Semantic Scene Understanding of LiDAR Sequences}, 
  year={2019}
}

@inproceedings{caesar2020nuscenes,
  author={Caesar, Holger and Bankiti, Varun and Lang, Alex H. and Vora, Sourabh and Liong, Venice Erin and Xu, Qiang and Krishnan, Anush and Pan, Yu and Baldan, Giancarlo and Beijbom, Oscar},
  booktitle={IEEE/CVF Conference on Computer Vision and Pattern Recognition (CVPR)}, 
  title={nuScenes: A Multimodal Dataset for Autonomous Driving}, 
  year={2020}
}

@inproceedings{cortinhal2020salsanext,
author="Cortinhal, Tiago
and Tzelepis, George
and Erdal Aksoy, Eren",
title="SalsaNext: Fast, Uncertainty-Aware Semantic Segmentation of LiDAR Point Clouds",
booktitle="Advances in Visual Computing",
year="2020"
}

@inproceedings{berman2018lovaszsoftmaxloss,
author={Berman, Maxim and Triki, Amal Rannen and Blaschko, Matthew B.},
booktitle={IEEE/CVF Conference on Computer Vision and Pattern Recognition (CVPR)}, 
title={The Lovasz-Softmax Loss: A Tractable Surrogate for the Optimization of the Intersection-Over-Union Measure in Neural Networks}, 
year={2018}
}

@INPROCEEDINGS{cheng2022cenet,
  author={Cheng, Hui–Xian and Han, Xian–Feng and Xiao, Guo–Qiang},
  booktitle={IEEE International Conference on Multimedia and Expo (ICME)}, 
  title={Cenet: Toward Concise and Efficient Lidar Semantic Segmentation for Autonomous Driving}, 
  year={2022}
}

@INPROCEEDINGS{wu2024ptv3,
  author={Wu, Xiaoyang and Jiang, Li and Wang, Peng-Shuai and Liu, Zhijian and Liu, Xihui and Qiao, Yu and Ouyang, Wanli and He, Tong and Zhao, Hengshuang},
  booktitle={IEEE/CVF Conference on Computer Vision and Pattern Recognition (CVPR)}, 
  title={Point Transformer V3: Simpler, Faster, Stronger}, 
  year={2024}
}

@ARTICLE{xu2025frnet,
  author={Xu, Xiang and Kong, Lingdong and Shuai, Hui and Liu, Qingshan},
  journal={IEEE Transactions on Image Processing}, 
  title={FRNet: Frustum-Range Networks for Scalable LiDAR Segmentation}, 
  year={2025},
  volume={34},
  number={},
  pages={2173-2186},
  doi={10.1109/TIP.2025.3550011}}

@inproceedings{bokhovkin2019boundaryloss,
author = {Bokhovkin, Alexey and Burnaev, Evgeny},
title = {Boundary Loss for Remote Sensing Imagery Semantic Segmentation},
year = {2019},
booktitle = {International Symposium on Neural Networks}
}

@inproceedings{abouhaidar2025harpnext,
  author    = {Abou Haidar, Samir and Chariot, Alexandre and Darouich, Mehdi and Joly, Cyril and Deschaud, Jean-Emmanuel},
  title     = {HARP-NeXt: High-Speed and Accurate Range-Point Fusion Network for 3D LiDAR Semantic Segmentation},
  booktitle = {Proceedings of the IEEE/RSJ International Conference on Intelligent Robots and Systems (IROS)},
  year      = {2025}
}

@inproceedings{xiao2022polarmix,
author = {Xiao, Aoran and Huang, Jiaxing and Guan, Dayan and Cui, Kaiwen and Lu, Shijian and Shao, Ling},
title = {PolarMix: a general data augmentation technique for LiDAR point clouds},
year = {2022},
isbn = {9781713871088},
publisher = {Curran Associates Inc.},
address = {Red Hook, NY, USA},
booktitle = {Proceedings of the 36th International Conference on Neural Information Processing Systems},
articleno = {802},
numpages = {14},
location = {New Orleans, LA, USA},
series = {NIPS '22}
}

@ARTICLE{sanchezcola2025,
  author={Sanchez, Jules and Deschaud, Jean-Emmanuel and Goulette, François},
  journal={IEEE Transactions on Robotics}, 
  title={COLA: COarse-LAbel Multisource LiDAR Semantic Segmentation for Autonomous Driving}, 
  year={2025},
  volume={41},
  number={},
  pages={1742-1754},
  doi={10.1109/TRO.2025.3543302}}

@ARTICLE{sanchez2024parisluco3d,
  author={Sanchez, Jules and Soum-Fontez, Louis and Deschaud, Jean-Emmanuel and Goulette, Francois},
  journal={IEEE Robotics and Automation Letters}, 
  title={ParisLuco3D: A High-Quality Target Dataset for Domain Generalization of LiDAR Perception}, 
  year={2024},
  volume={9},
  number={6},
  pages={5496-5503},
  doi={10.1109/LRA.2024.3393209}}

@inproceedings{li2018learningtogeneralize,
  title={Learning to generalize: Meta-learning for domain generalization},
  author={Li, Da and Yang, Yongxin and Song, Yi-Zhe and Hospedales, Timothy},
  booktitle={Proceedings of the AAAI conference on artificial intelligence},
  volume={32},
  number={1},
  year={2018}
}

@INPROCEEDINGS{yi2021domainadaptation,
  author={Yi, Li and Gong, Boqing and Funkhouser, Thomas},
  booktitle={2021 IEEE/CVF Conference on Computer Vision and Pattern Recognition (CVPR)}, 
  title={Complete and Label: A Domain Adaptation Approach to Semantic Segmentation of LiDAR Point Clouds}, 
  year={2021},
  volume={},
  number={},
  pages={15358-15368},
  doi={10.1109/CVPR46437.2021.01511}}

@INPROCEEDINGS{sanchez2023domaingeneralization,
  author={Sanchez, Jules and Deschaud, Jean-Emmanuel and Goulette, François},
  booktitle={2023 IEEE/CVF International Conference on Computer Vision (ICCV)}, 
  title={Domain generalization of 3D semantic segmentation in autonomous driving}, 
  year={2023},
  volume={},
  number={},
  pages={18031-18041},
  doi={10.1109/ICCV51070.2023.01657}}

@article{sanchez20253dlabelprop,
  title={3DLabelProp: Geometric-Driven Domain Generalization for LiDAR Semantic Segmentation in Autonomous Driving},
  author={Sanchez, Jules and Deschaud, Jean-Emmanuel and Goulette, François},
  journal={arXiv preprint arXiv:250- 1.14605},
  year={2025}
}

@INPROCEEDINGS{li2019episodictraining,
  author={Li, Da and Zhang, Jianshu and Yang, Yongxin and Liu, Cong and Song, Yi-Zhe and Hospedales, Timothy},
  booktitle={2019 IEEE/CVF International Conference on Computer Vision (ICCV)}, 
  title={Episodic Training for Domain Generalization}, 
  year={2019},
  volume={},
  number={},
  pages={1446-1455},
  doi={10.1109/ICCV.2019.00153}}

@ARTICLE{jin2022stylenormalization,
  author={Jin, Xin and Lan, Cuiling and Zeng, Wenjun and Chen, Zhibo},
  journal={IEEE Transactions on Multimedia}, 
  title={Style Normalization and Restitution for Domain Generalization and Adaptation}, 
  year={2022},
  volume={24},
  number={},
  pages={3636-3651},
  doi={10.1109/TMM.2021.3104379}}

@INPROCEEDINGS{langer2020domaintransfer,
  author={Langer, Ferdinand and Milioto, Andres and Haag, Alexandre and Behley, Jens and Stachniss, Cyrill},
  booktitle={2020 IEEE/RSJ International Conference on Intelligent Robots and Systems (IROS)}, 
  title={Domain Transfer for Semantic Segmentation of LiDAR Data using Deep Neural Networks}, 
  year={2020},
  volume={},
  number={},
  pages={8263-8270},
  doi={10.1109/IROS45743.2020.9341508}}

@INPROCEEDINGS{kong2023robo3d,
  author={Kong, Lingdong and Liu, Youquan and Li, Xin and Chen, Runnan and Zhang, Wenwei and Ren, Jiawei and Pan, Liang and Chen, Kai and Liu, Ziwei},
  booktitle={2023 IEEE/CVF International Conference on Computer Vision (ICCV)}, 
  title={Robo3D: Towards Robust and Reliable 3D Perception against Corruptions}, 
  year={2023},
  volume={},
  number={},
  pages={19937-19949},
  doi={10.1109/ICCV51070.2023.01830}}

@inproceedings{
dan2019benchmarkingrobustness,
title={Benchmarking Neural Network Robustness to Common Corruptions and Perturbations},
author={Dan Hendrycks and Thomas Dietterich},
booktitle={International Conference on Learning Representations},
year={2019}
}
% \bibliography{IEEEabrv, iros_main}

\end{document}